\documentclass[10pt,twocolumn,letterpaper]{article}

\usepackage[pagenumbers]{cvpr} %

\definecolor{cvprblue}{rgb}{0.21,0.49,0.74}
\usepackage[pagebackref,breaklinks,colorlinks,allcolors=cvprblue]{hyperref}
\usepackage{graphicx}
\usepackage{array}
\usepackage{multirow}
\usepackage[margin=1in]{geometry}
\usepackage{caption}
\usepackage{booktabs}
\usepackage{tabularx}
\usepackage{multicol}
\usepackage{enumitem}
\usepackage{subcaption}
\usepackage{makecell}
\newcolumntype{C}[1]{>{\centering\arraybackslash}p{#1}}
\usepackage{amsmath, amssymb}
\usepackage[many]{tcolorbox}
\usepackage[utf8]{inputenc}
\usepackage[T1]{fontenc}
\usepackage[ruled,linesnumbered]{algorithm2e}
\usepackage{algpseudocode}

\def\paperID{37632} %
\def\confName{CVPR}
\def\confYear{2026}
\title{Cross-Resolution Diffusion Models via Network Pruning
}

\author{
Jiaxuan Ren$^{2*\ddagger}$ \quad 
Junhan Zhu$^{1*}$ \quad 
Huan Wang$^{1\dagger}$ \\
$^{1}$Westlake University \quad 
$^{2}$University of Electronic Science and Technology of China
\\
\textcolor{magenta}{\texttt{\url{https://xuan9-9.github.io/CR-Diff/}}}
}

\begin{document}
\twocolumn[{%
\renewcommand\twocolumn[1][]{#1}%
\maketitle
\begin{center}
    \centering
    \vspace{-8pt}
    \captionsetup{font={small}, skip=8pt}
    \includegraphics[width=1\linewidth]{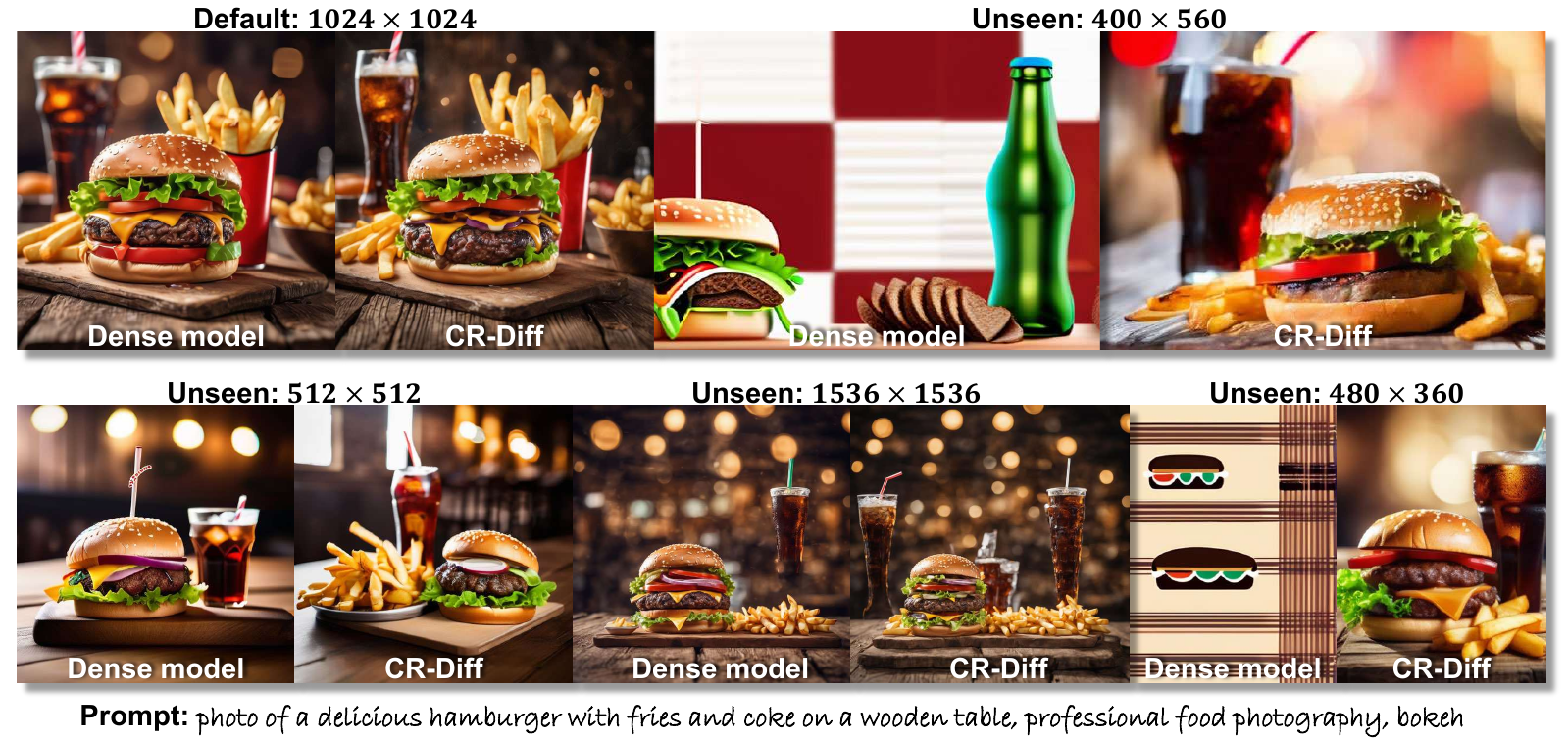}
    \vspace{-15pt}
    \captionof{figure}{This paper presents \textit{CR-Diff}, a method to improve the \underline{c}ross-\underline{r}esolution visual consistency of UNet--based \underline{diff}usion models by masking out some parameters in the model, \textit{i.e.}, \textit{network pruning} -- a technique that has been widely used for reducing model size; while here, we novelly repurpose it for generalizing diffusion models to unseen resolutions. The samples above compare the original SDXL~\citep{podell2023sdxl} model with its counterpart modified by our proposed CR-Diff.  The original SDXL is trained at 1024$\times$1024 resolution and can hardly generalize to other resolutions (\textit{e.g.}, 400$\times$560, 480$\times$360), while after CR-Diff prunes some parameters (the kept parameters are unchanged), it manages to generate much more coherent images at these unseen resolutions. This phenomenon suggests some parameters in the UNet-based diffusion models may be like a kind of ``impurity''; while pruning, which used to be deemed to \textit{damage} the model's capacity, can actually ``purify'' the diffusion model, improving its generalizability across resolutions.
    }
\label{fig:teaser}
\vspace{15pt}
\end{center}%
}]

\vspace{-1em}
\renewcommand{\thefootnote}{\fnsymbol{footnote}}
\makeatletter
\renewcommand{\@makefntext}[1]{%
  \parindent 1em\noindent
  \hbox to 1.8em{#1\hss}} %
\makeatother
\footnotetext[1]{$^*$These authors contributed equally to this work.}
\footnotetext[2]{$^\dagger$Corresponding author: \texttt{wanghuan@westlake.edu.cn}}
\footnotetext[3]{%
$^\ddagger$\parbox[t]{1.0\linewidth}{
 Work done as a visiting research intern at ENCODE Lab, Westlake University.
}}
\begin{abstract}
Diffusion models have demonstrated impressive image synthesis performance, yet many UNet–based models are trained at certain fixed resolutions. Their quality tends to degrade when generating images at out-of-training resolutions.
We trace this issue to resolution-dependent parameter behaviors, where weights that function well at the default resolution can become adverse when spatial scales shift, weakening semantic alignment and causing structural instability in the UNet architecture.
Based on this analysis, this paper introduces \textbf{CR-Diff}, a novel method that improves the cross-resolution visual consistency by pruning some parameters of the diffusion model.
Specifically, CR-Diff has two stages.
It first performs block-wise pruning to selectively eliminate adverse weights. Then, a pruned output amplification is conducted to further purify the pruned predictions.
Empirically, extensive experiments suggest that CR-Diff can improve perceptual fidelity and semantic coherence across various diffusion backbones and unseen resolutions, while largely preserving the performance at default resolutions. 
Additionally, CR-Diff supports prompt-specific refinement, enabling quality enhancement on demand.
\end{abstract}

\vspace{-20pt}
\section{Introduction}
\label{sec:intro}

Diffusion models~\citep{sohl2015deep,ho2020denoising,songscore,song2019generative,songdenoising,dhariwal2021diffusion} have achieved remarkable success in text-to-image generation~\citep{saharia2022photorealistic,nichol2022glide,rombach2022high,podell2023sdxl,esser2024scaling,xie2024sana}, enabling high-quality synthesis across a wide range of visual concepts.
However, despite their strong generative capacity, most models are trained at \textit{default resolutions}~(e.g., $1024 \times 1024$ for SDXL~\citep{podell2023sdxl}).
Although techniques like multi-aspect bucket sampling~\citep{podell2023sdxl, novelai2022improvements} provide some flexibility by fine-tuning on various aspect ratios, the core problem persists.
When applied to \textit{unseen resolutions} outside the training regime, these models tend to exhibit obvious artifacts, reduced semantic alignment, and diminished structural coherence.
Recent DiT-based models~\citep{esser2024scaling,BlackForestLabs2024Flux} natively address this limitation through scale-adaptive position encodings.
In contrast, foundational UNet-based~\citep{ronneberger2015u} models~\citep{rombach2022high} lack such inherent robustness, making their generative quality more sensitive to changes in spatial scale.

\begin{figure}[htbp]
    \centering
    \captionsetup[subfigure]{font=scriptsize}
    \begin{subfigure}[t]{1\linewidth}
        \centering
        \includegraphics[width=1\linewidth]{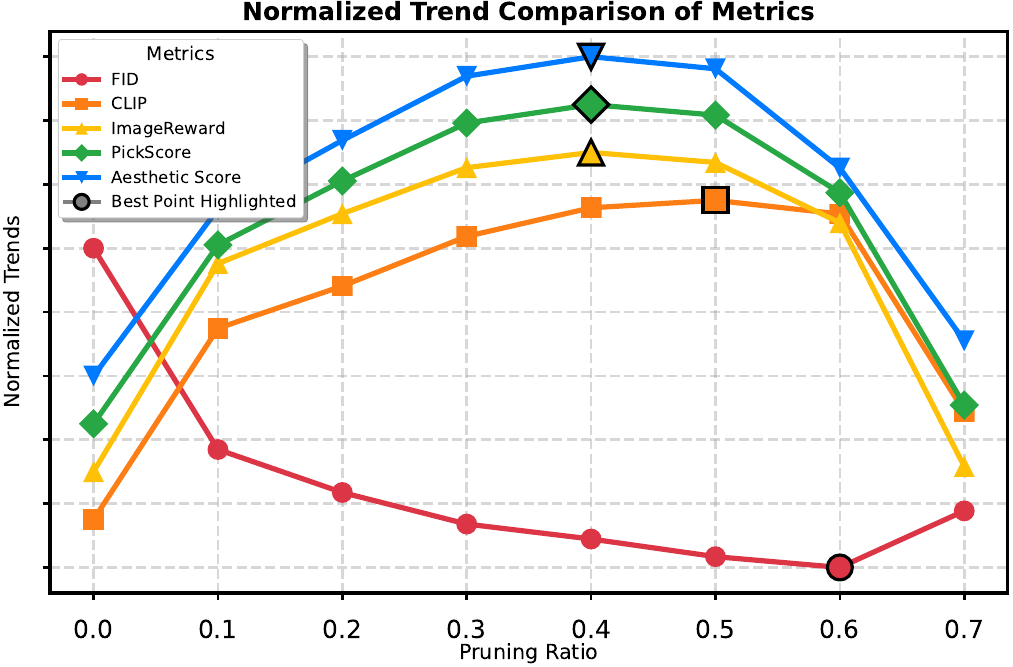} %
        \caption{}
        \label{fig:motivationa}
    \end{subfigure}
    
    \vspace{5pt} %

    \begin{subfigure}[t]{1\linewidth}
        \centering
        \includegraphics[width=1\linewidth]{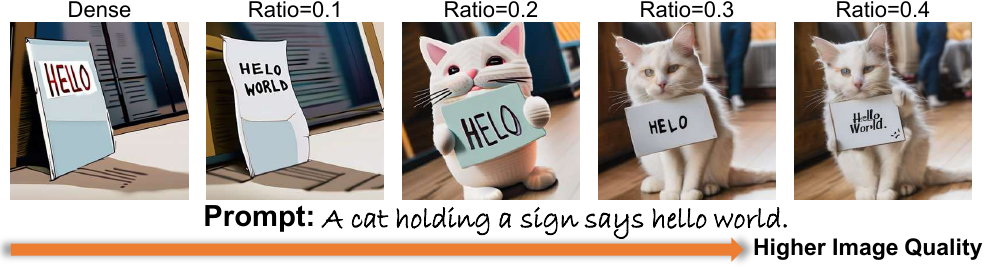} %
        \caption{}
        \label{fig:motivationb}
    \end{subfigure}
    \vspace{-6pt}
    \caption{Effects of magnitude-based unstructured pruning on SDXL at unseen resolution $512\times512$. (a) Quantitative metrics improvement within moderate sparsity. (b) Qualitative illustration of improved semantic alignment as sparsity increases. 
    } %
    \label{fig:motivation}
    \vspace{-15pt}
\end{figure}

Network pruning~\citep{han2015learning,han2016deep,wang2021neural,fang2023depgraph,feng2024oracle,wang2022trainability} is traditionally used to improve efficiency by reducing computation and memory cost~\citep{fang2305structural,fang2025tinyfusion,li2023snapfusion,zhao2024mobilediffusion,castells2024ld}. These approaches primarily aim to compress models while preserving accuracy. Surprisingly, here we observe that pruning in diffusion UNets can play a qualitatively different role. As shown in Figure~\ref{fig:motivation}, when applying simple magnitude pruning to SDXL at the unseen resolution of $512\times512$, we observe a counter-intuitive trend. Instead of degrading performance, moderate sparsity \textit{improves} generation quality. In Figure~\ref{fig:motivationa}, metrics such as ImageReward steadily increase as sparsity rises from 0\% to 40\%, while FID decreases accordingly. This quantitative gain is further reflected in the visual samples in Figure~\ref{fig:motivationb}. At 0\% sparsity, the dense model fails to produce a coherent object~(the “cat” is missing, and the text is incomplete). As sparsity increases to 10–30\%, the generated content becomes more semantically aligned. At 40\%, both the concept of “a cat holding a sign” and the phrase “hello world” are rendered clearly. 

Such phenomena suggest that parameters beneficial at the default resolutions can become \textit{adverse} when applied to unseen resolutions, and pruning mitigates these effects and helps stabilize the generative process. All of these observations lead us to ask: \textit{Can we devise a controllable pruning-based strategy to improve the cross-resolution generability of UNet-based diffusion models?}

\begin{figure*}
\centering
\captionsetup{font={small}, skip=8pt}
\includegraphics[width=1\linewidth]{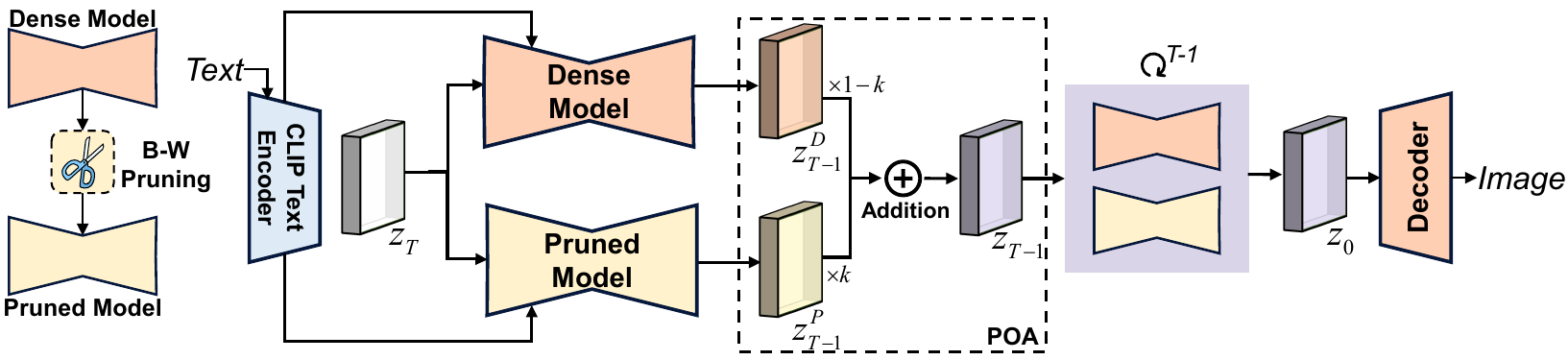}
\vspace{-15pt}
\caption{\textbf{Overview of CR-Diff.} Most UNet–based diffusion models exhibit resolution-dependent degradation when generating at unseen scales. CR-Diff addresses this issue through a two-stage \textit{pruning and optimizing} process, consisting of a \textbf{block-wise~(B-W) pruning ratio} strategy and a \textbf{pruned output amplification~(POA)} mechanism. As shown in Figure~\ref{fig:block-wise}, \textit{block-wise pruning} adopts a magnitude-based criterion with adaptive ratios across blocks to extract parameter directions that remain stable across resolutions.
\textit{Pruned output amplification} refines the model’s forward predictions by amplifying the pruned output with an amplification coefficient $k>1$, which suppresses residual dense model's influences that otherwise introduce artifacts.
This leads to cleaner denoising trajectories and higher-quality final images with more stable structure and details.
} 
\vspace{-12pt}
\label{fig:main}
\end{figure*}

To this end, we introduce \textit{CR-Diff}, a two-stage framework that restructures parameter distribution and purifies model outputs of diffusion UNets for improved generation quality at unseen resolutions while maintaining performance at default ones. 
As shown in Figure~\ref{fig:main}, CR-Diff first applies \textit{block-wise pruning} to assign differentiated pruning ratios across downsampling, middle, and upsampling blocks, yielding a pruned backbone reflecting the intrinsic importance distribution. Then, a \textit{pruned output amplification} mechanism further purifies predictions by rebalancing dense and pruned outputs, enhancing beneficial signals while suppressing adverse ones. CR-Diff further supports \textit{prompt-specific refinement}, allowing targeted quality enhancement. All are achieved without altering the model architecture, remaining effective across resolutions as shown in Figure~\ref{fig:teaser}.

Our contributions are summarized as follows:
\begin{itemize}[leftmargin=*]
\item We reveal that pruning diffusion UNets can \textit{improve} text-to-image performance, particularly at unseen resolutions where dense models exhibit resolution bias.
\item We introduce a \textit{block-wise pruning} and \textit{output amplification} strategy that adapts sparsity across the UNet and refines the pruned subnetwork to improve generation quality and stabilize semantic coherence.
\item Experiments show that our method consistently and controllably enhances output quality, improving various metrics across models and resolutions.
\end{itemize}
\vspace{5pt}

\section{Related Work}

\paragraph{Text-to-Image Diffusion Models.} Diffusion models~\citep{ho2020denoising,songscore} have established themselves as the state-of-the-art for high-fidelity text-to-image synthesis, powering models like the widely-used Stable Diffusion~(SD) series~\citep{rombach2022high,podell2023sdxl,salimans2022progressive,esser2024scaling}, DALL-E2~\citep{ramesh2022hierarchical}, sana~\citep{xiesana,xie2024sana,chen2025sana3}, Pixart~\citep{chen2024pixart,chen2024pixart2,chen2024pixart3}, and FLUX~\citep{BlackForestLabs2024Flux}. However, a significant limitation of traditional UNet architectures, particularly foundational models like SD 1.5, is their limited generalization to resolutions and aspect ratios unseen during training. This fragility largely stems from spatially fixed inductive biases such as learned positional encodings in attention layers. Consequently, generating images at novel resolutions directly often leads to obvious degradation in visual coherence and semantic fidelity, such as object duplication or compositional collapse.

To mitigate this, several strategies have been proposed. The most common approach is multi-aspect training~\citep{podell2023sdxl,novelai2022improvements}, where models are explicitly fine-tuned on data "bucketed" into various aspect ratios after pretraining models at a fixed aspect-ratio and resolution, as was done for SDXL~\citep{podell2023sdxl}.  More recently, MMDiT-based~\citep{peebles2023scalable} architectures like SD3~\citep{esser2024scaling} and FLUX~\citep{BlackForestLabs2024Flux} have demonstrated superior flexibility by design. Instead of interpolating fixed embeddings~\citep{dosovitskiy2020image}, they natively handle variable input dimensions by generating 2D positional grids, which are constructed based on maximum training dimensions and then center-cropped to target resolutions before being frequency embedded. In contrast to these approaches, our work introduces a novel method to improve generation quality at unseen resolutions through a post-hoc, pruning-based strategy.

\vspace{-15pt}
\paragraph{Neural Network Pruning.}
Neural network pruning~\citep{lecun1989optimal,hassibi1993optimal,han2015learning,han2016deep,wang2021neural,aghasi2017net,feng2024oracle,fang2023depgraph} is widely used to reduce parameter count and computational cost in deep learning, and has recently seen applications in large language models~\citep{frantar2023sparsegpt,sunsimple,ling2024slimgpt,wei2024structured} as well as other large-scale architectures~\citep{tuo2025sparsessm,shen2025sparse,shihab-etal-2025-efficient,frantar2022optimal}. 
In diffusion models, pruning has primarily been explored as a compression technique to improve inference efficiency, leading to compact generators such as SnapFusion~\citep{li2023snapfusion}, MobileDiffusion~\citep{zhao2024mobilediffusion}, BK-SDM~\citep{kim2024bk}, Laptop-Diff~\citep{zhang2024laptop}, and LD-Pruner~\citep{castells2024ld}. Recent general-purpose frameworks, including EcoDiff~\citep{zhang2024effortless} and OBS-Diff~\citep{zhu2025obs}, also follow this compression-oriented objective.

However, these methods view pruning solely as a means of model compression. In contrast, we find that pruning can \textit{improve} the generative quality of text-to-image diffusion models, revealing a qualitatively different role for sparsity beyond efficiency.
\label{sec:related}

\section{Method: CR-Diff}
\label{sec:method}

\subsection{Preliminaries}
\label{sec:preliminaries}
Diffusion models~\citep{ho2020denoising,rombach2022high} generate images by progressively denoising a latent variable $x_t$ through a learned reverse diffusion process parameterized by a UNet backbone. Given a noisy latent $x_t$ at timestep $t$, the model predicts the clean signal $\hat{x}_0$ conditioned on a text or image prompt $c$. The training objective is formulated as:
\begin{equation}
    \mathcal{L} = \mathbb{E}_{x_0, t, \epsilon}\left[\|\epsilon - \epsilon_\theta(x_t, t, c)\|_2^2\right],
    \label{eq:diffusion}
\end{equation}
where $\epsilon \sim \mathcal{N}(0, I)$ and $x_t = \sqrt{\bar{\alpha}_t}x_0 + \sqrt{1 - \bar{\alpha}_t}\epsilon$. The denoising network $\epsilon_\theta$ is realized as a hierarchical UNet consisting of convolutional and attention-based modules distributed across multiple feature resolutions. 
\paragraph{Block-Specific Contribution Pattern.}
The UNet architecture in diffusion models can be decomposed into three structural stages, namely the \textit{downsampling blocks}, the \textit{middle blocks}, and the \textit{upsampling blocks}. These stages operate at distinct feature scales and serve complementary purposes in the generative process. These functional asymmetries cause different blocks to contribute unevenly to the denoising process. The ablation results in Table~\ref{block-wise-table} prove that optimal pruning ratios vary accordingly, and \textit{applying differentiated treatment across blocks} leads to improved performance.
 
\vspace{-5pt}
\paragraph{Resolution-Sensitive Weight Behavior.}
Diffusion UNets comprise convolution layers that capture local spatial priors and fine-grained textures, attention layers that establish global semantic relationships and text–image alignment, feed-forward layers that reshape intermediate feature representations, and normalization or modulation parameters that encode activation statistics across diffusion steps. Although jointly trained, these components are implicitly adapted to the feature and scale statistics of the trained default resolution. 

Consequently, when the model is applied to \textit{unseen resolutions}, the feature distributions shift away from those seen during training. Scale-specific weights no longer align with the altered structure. In such cases, these parameters can be regarded collectively as \textit{adverse weights}, referring to weights that do not align well with the semantic structure required at non-default resolutions, and can lead to degraded visual coherence when generating at unseen resolutions. 

\subsection{Overall Framework}
\label{sec:overall}

\begin{figure}
\centering
\captionsetup{font={small}, skip=8pt}
\includegraphics[width=1\linewidth]{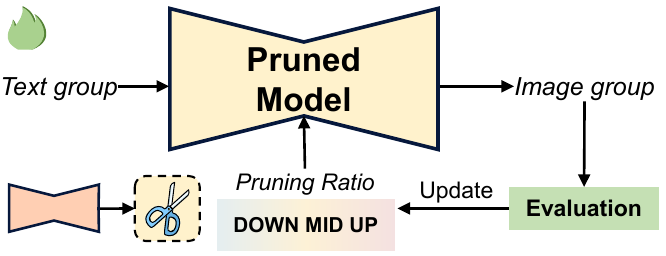}
\vspace{-13pt}
\caption{
Simulated annealing~(SA) search process for determining the optimal block-wise pruning ratios.
}
\label{fig:SA}
\vspace{-18pt}
\end{figure}

Building upon the diffusion UNet foundation introduced above, our pruning framework seeks to preserve semantically essential parameters while attenuating adverse ones, thereby improving image generation. 
The central idea is to apply block-wise sparsification across the UNet hierarchy and subsequently refine the retained subnetwork to mitigate residual degradation.
As illustrated in Figure~\ref{fig:main}, the framework operates in two sequential stages, \textit{pruning} and \textit{optimization}. 

\begin{figure}[t]
    \centering
\captionsetup{font={small}, skip=8pt}
\includegraphics[width=1\linewidth]{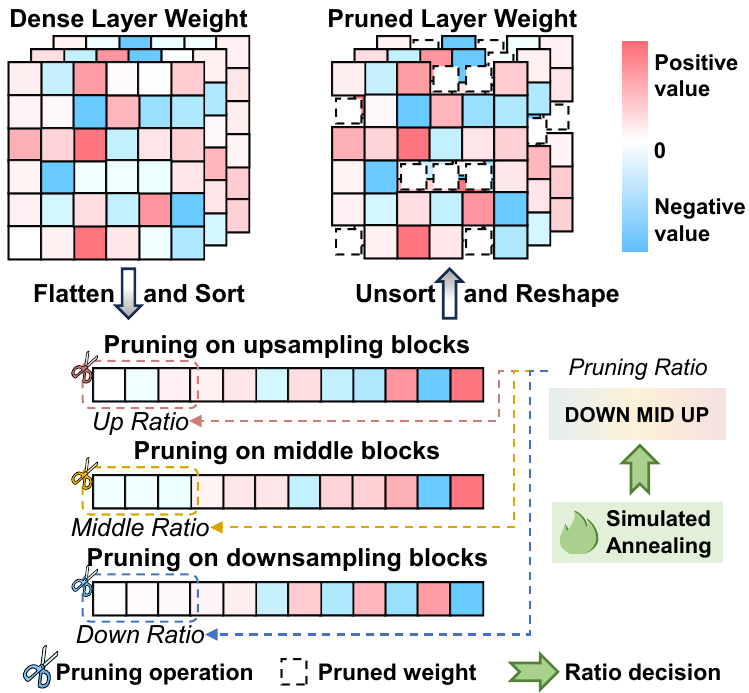}
\vspace{-13pt}
\caption{Block-wise~(B-W) pruning applies differentiated pruning ratios across blocks to preserve essential structure and provide a pruned backbone for subsequent optimization.}
    \label{fig:block-wise}
    \vspace{-13pt}
\end{figure}

In the \textit{pruning} stage, the \textbf{block-wise pruning ratio} strategy shown in Figure~\ref{fig:block-wise} assigns differentiated pruning ratios to the downsampling, middle, and upsampling blocks, which improves generation quality and yields a pruned backbone that reflects intrinsic importance distributions of weights.
In the \textit{optimization} stage, the \textbf{pruned output amplification} (POA) mechanism shown in Figure~\ref{fig:main} leverages differences between dense and pruned outputs, amplifying pruned prediction while attenuating residual dense signals that introduce artifacts.

After the two-stage refinement, CR-Diff can synthesize images from text prompts with cleaner denoising trajectories and noticeably improved visual quality.

\subsection{Block-Wise Pruning Ratio Strategy}
\label{sec:blockwise}

\begin{table*}[t]
\centering
\caption{
Performance comparison on unseen resolutions. Across the evaluated models and resolutions, CR-Diff improves most metrics relative to the dense model, with \textbf{bold} values indicating superior performance of our CR-Diff against the dense model.
}
\vspace{-5pt}
\label{table:AG_unseen}
\resizebox{\textwidth}{!}{%
 \begin{tabular}{l c cc cc cc cc cc}
 \toprule
 \multirow{2}{*}{Model} & \multirow{2}{*}{Resolution} & \multicolumn{2}{c}{FID $\downarrow$} & \multicolumn{2}{c}{CLIP $\uparrow$} & \multicolumn{2}{c}{ImageReward $\uparrow$}& \multicolumn{2}{c}{PickScore $\uparrow$}& \multicolumn{2}{c}{Aesthetic Score $\uparrow$} \\
 \cmidrule(r){3-4} \cmidrule(lr){5-6} \cmidrule(lr){7-8} \cmidrule(lr){9-10} \cmidrule(l){11-12}
 & & \multicolumn{1}{c}{Dense} & \multicolumn{1}{c}{CR-Diff} & \multicolumn{1}{c}{Dense} & \multicolumn{1}{c}{CR-Diff} & \multicolumn{1}{c}{Dense} & \multicolumn{1}{c}{CR-Diff} & \multicolumn{1}{c}{Dense} & \multicolumn{1}{c}{CR-Diff} & \multicolumn{1}{c}{Dense} & \multicolumn{1}{c}{CR-Diff} \\
 \midrule
 \multirow{3}{*}{SDXL} & $512 \times 512$ & 83.827 & \textbf{37.918} & 0.295 & \textbf{0.321} & -0.498 & \textbf{0.735} & 20.296 & \textbf{22.140} & 4.335 & \textbf{5.525} \\
 & $ 400\times560 $ & 146.984 & \textbf{36.688} & 0.252 & \textbf{0.311} & -1.734 & \textbf{0.092} & 18.608 & \textbf{21.074} & 3.494 & \textbf{4.672} \\
 & $ 480\times360 $ & 211.369 & \textbf{46.040} & 0.225 & \textbf{0.307} & -2.148 & \textbf{-0.099} & 18.060 & \textbf{20.956} & 3.806 & \textbf{4.644} \\
\midrule
\multirow{3}{*}{SD1.5} & $ 400\times560 $ & 39.047 & 39.291 & 0.309 & \textbf{0.310} & 0.061 & \textbf{0.151} & 21.146 & \textbf{21.188} & 4.736 & \textbf{4.779} \\
& $ 480\times360 $ & 39.797 & \textbf{37.634} & 0.307 & \textbf{0.307} & -0.068 & \textbf{-0.026} & 20.906 & \textbf{20.944} & 4.710 & \textbf{4.819} \\
& $ 768\times768 $ & 38.832 & \textbf{38.452} & 0.314 & \textbf{0.315} & -0.050 & \textbf{0.059} & 21.208 & \textbf{21.232} & 5.419 & 5.385 \\
\midrule
\multirow{3}{*}{SD2.1} & $ 400\times560 $ & 48.110 & \textbf{35.837} & 0.296 & \textbf{0.304} & -0.461 & \textbf{-0.068} & 20.374 & \textbf{20.540} & 4.190 & \textbf{4.428} \\
 & $ 480\times360 $ & 73.807 & \textbf{41.042} & 0.278 & \textbf{0.294} & -0.933 & \textbf{-0.561} & 19.573 & \textbf{20.177} & 3.984 & \textbf{4.532} \\
& $ 768\times768 $ & 37.198 & \textbf{35.237} & 0.317 & \textbf{0.318} & 0.334 & \textbf{0.419} & 21.695 & 21.451 & 5.583 & 5.339 \\
\bottomrule
\end{tabular}%
}
\vspace{-8pt}
\end{table*}

As discussed in Section~\ref{sec:preliminaries}, heterogeneous weight functions make uniform pruning ratios less effective on diffusion UNets, which inevitably reduces local texture encoding in early downsampling blocks, diminishes global semantic integration in middle blocks, and limits high-frequency detail recovery in upsampling blocks, ultimately compromising visual coherence and fidelity. 

To this end, we employ a \textit{block-wise pruning ratio} strategy, in which downsampling blocks, the middle block, and upsampling blocks of the UNet are each assigned distinct pruning ratios based on magnitude.

To determine the optimal pruning ratio for each block, we adopt a \textit{simulated annealing~(SA)} search strategy. Let $\mathbf{r} = { r_{\text{down}}, r_{\text{mid}}, r_{\text{up}} }$ denote the pruning ratio configuration across the downsampling, middle, and upsampling blocks. Starting from initial configurations, the model generates images for a fixed set of prompts, and their ImageReward is averaged to assess the overall performance of the current ratio setting. SA then perturbs and updates $\mathbf{r}$ iteratively, gradually refining it to maximize generation quality. The search procedure is illustrated in Figure~\ref{fig:SA}, and the full algorithm is provided in the supplementary material due to limited space.

Through this exploration, each block receives a ratio that preserves critical semantic structure while suppressing weights that introduce degradation in generation.

\vspace{3pt}

\subsection{Pruned Output Amplification}
\label{sec:amplification}

To further refine the generative behavior of the pruned model, we introduce a \textit{pruned output amplification}~(POA) mechanism, which operates on the forward denoising trajectory, as illustrated in Figure~\ref{fig:main}. At each denoising step $t$, we obtain the predicted output $\mathbf{z}_{t}^P$ from the pruned model and the corresponding output $\mathbf{z}_{t}^D$ from the dense model, and then performs combination:
\begin{equation}
\mathbf{z}_{t} = k\,\mathbf{z}_{t}^P + (1 - k)\,\mathbf{z}_{t}^D,
\label{eq:poa}
\end{equation}
where the amplification coefficient $k$ determines the relative contribution of the pruned and dense outputs.

Because $\mathbf{z}_{t}^P - \mathbf{z}_{t}^D$ represents the pruning-induced shift that improves generative behavior, choosing $k>1$ selectively amplifies this beneficial direction while suppressing residual artifact-inducing tendencies inherited from the dense model. This step-by-step refinement stabilizes the denoising trajectory and preserves structural consistency throughout sampling. After applying both block-wise pruning and POA, the resulting model produces higher-quality images directly from text prompts.

\section{Experiments}
\label{sec:exp}

\subsection{Settings}
\label{sec:Settings}
\paragraph{Models and Resolutions.}
To assess the effectiveness of CR-Diff, we apply pruning to three UNet–based diffusion models across both their default training resolutions and a set of unseen resolutions. For SDXL~\citep{podell2023sdxl}, in addition to its default $1024 \times 1024$ resolution, we evaluate performance at $512 \times 512$, $400 \times 560$, and $480 \times 360$. For SD1.5 and SD2.1~\citep{rombach2022high}, beyond the default $512 \times 512$, we likewise consider $400 \times 560$, $480 \times 360$ ,and $768 \times 768$ as unseen settings. This setup enables us to examine how pruning influences generative robustness when moving away from the resolution regime on which the model was originally trained. In the following experiments, all resolutions are expressed in the format height $\times$ width.

\begin{figure*}
\centering
\captionsetup{font={small}, skip=8pt}
\includegraphics[width=1\linewidth]{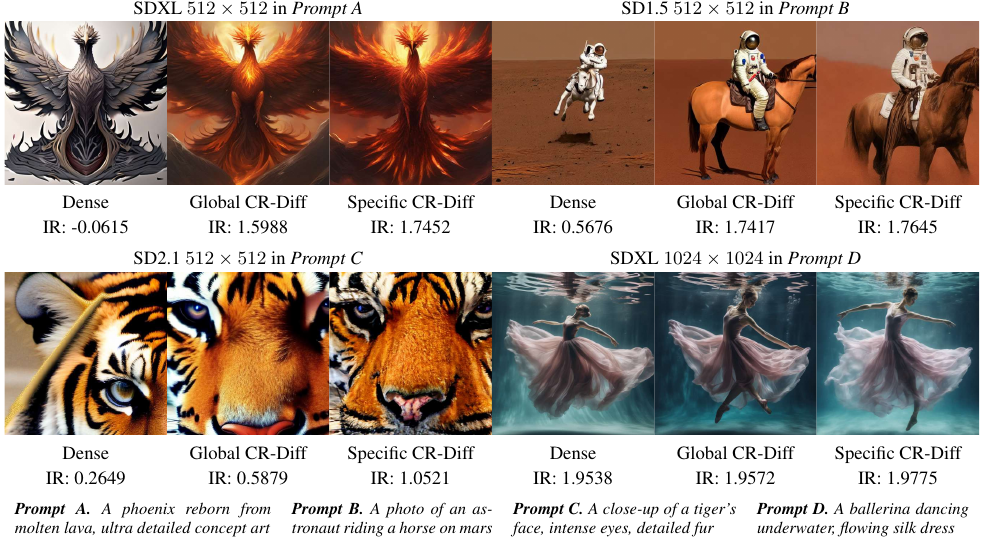}
\vspace{-15pt}
\caption{Visual comparison across three generation settings. 
    \textit{Dense} denotes the original unpruned model. 
    \textit{Global CR-Diff} applies a pruning ratio optimized for a specific model and resolution, shared across all prompts. 
    \textit{Specific CR-Diff} further refines this ratio for the given prompt, enabling prompt-specific optimization. 
    Each group corresponds to a specific prompt, and the ImageReward (IR) scores are shown below each image. 
    Global CR-Diff improves generative fidelity in a prompt-agnostic manner, while Specific CR-Diff further enhances semantic alignment and visual coherence for the specific prompt.}
    \vspace{-5pt}
    \label{fig:single}
\end{figure*}

\vspace{-5pt}
\paragraph{Evaluation Metrics.}
We evaluate our method on a subset of 5K prompts sampled from the MS-COCO 2014 validation set~\citep{lin2014microsoft}. Performance is measured along three dimensions: image fidelity, text–image alignment, and aesthetic preference. Specifically, Fréchet Inception Distance (FID)~\citep{heusel2017gans} is used to assess image quality, while CLIP Score~\citep{hessel2021clipscore} and ImageReward~\citep{xu2023imagereward} evaluate semantic alignment between text and image. Plus, PickScore~\citep{kirstain2023pick} and Aesthetic Score provide assessments of aesthetic appeal and human preference consistency.

\subsection{Results of CR-Diff on Unseen Resolutions}
\label{sec:unseen}
As shown in Table~\ref{table:AG_unseen}, CR-Diff demonstrates generally improved performance across multiple diffusion backbones when evaluated at resolutions that deviate from their default training settings. 

For SDXL, which is originally optimized for the resolution at $1024 \times 1024$, applying CR-Diff at unseen resolutions results in substantial and notable gains across all evaluation metrics. 
The large magnitude of improvement suggests that scale-mismatched parameters in SDXL strongly contribute to texture degradation and structural inconsistency, and that CR-Diff effectively suppresses these detrimental effects. 

For SD1.5 and SD2.1, which are natively trained at $512 \times 512$, CR-Diff also provides consistent gains when evaluated at unseen resolutions. Improvements are reflected in enhanced semantic alignment as measured by CLIP and ImageReward, as well as better visual preference captured by PickScore and Aesthetic Score. 

Compared with SDXL, however, the improvements appear more moderate. This is due to the intrinsic resolution characteristics of SD1.5 and SD2.1. Their training data encourages coarser semantic representation, with objects occupying larger spatial regions and containing relatively low detail density. As a result, reducing resolution does not heavily disrupt global structure because the models are designed to perform well under limited texture complexity.

\begin{table*}[t]
\centering
\caption{
Performance comparison on default resolutions.  
Across the evaluated models, CR-Diff consistently improves or maintains performance, with \textbf{bold} values indicating gains over the dense model.
}
\vspace{-5pt}
\label{table:AG_default} 
\resizebox{\textwidth}{!}{
\begin{tabular}{l c cc cc cc cc cc}
\toprule
\multirow{2}{*}{Model} & \multirow{2}{*}{Resolution} & \multicolumn{2}{c}{FID $\downarrow$} & \multicolumn{2}{c}{CLIP $\uparrow$} & \multicolumn{2}{c}{ImageReward$\uparrow$}& \multicolumn{2}{c}{PickScore $\uparrow$}& \multicolumn{2}{c}{Aesthetic Score $\uparrow$} \\
\cmidrule(r){3-4} \cmidrule(lr){5-6} \cmidrule(lr){7-8} \cmidrule(lr){9-10} \cmidrule(l){11-12}
& & \multicolumn{1}{c}{Dense} & \multicolumn{1}{c}{CR-Diff} & \multicolumn{1}{c}{Dense} & \multicolumn{1}{c}{CR-Diff} & \multicolumn{1}{c}{Dense} & \multicolumn{1}{c}{CR-Diff} & \multicolumn{1}{c}{Dense} & \multicolumn{1}{c}{CR-Diff} & \multicolumn{1}{c}{Dense} & \multicolumn{1}{c}{CR-Diff} \\
\midrule
SDXL & $1024 \times 1024$ & 33.186 & 33.562 & 0.322 & \textbf{0.322} & 0.788 & \textbf{0.946} & 22.512 & \textbf{22.639} & 6.123 & 6.106 \\
SD1.5& $ 512\times512 $ & 38.368 & \textbf{37.773} & 0.315 & 0.314 & 0.239 & 0.203 & 21.539 & 21.377 & 5.205 & \textbf{5.233} \\
SD2.1& $ 512\times512 $ & 45.583 & \textbf{36.792} & 0.308 & \textbf{0.309} & -0.100 & \textbf{-0.052} & 20.943 & \textbf{20.960} & 4.728 & \textbf{5.082} \\
\bottomrule
\end{tabular}
}
\vspace{-15pt}
\end{table*}

\subsection{Results of Prompt-Specific Optimization}
\label{sec:single}
CR-Diff already provides substantial gains over the dense model, with improvements observed on over 85\% of evaluated prompts under global refinement. This demonstrates that the two-stage framework is broadly effective in enhancing overall fidelity and semantic consistency across diverse scenes. Nevertheless, some prompts involve particularly fine-grained textures, rare materials, or compositionally intricate structures that can benefit from more specialized treatment than what global refinement alone can supply. For such cases, CR-Diff provides prompt-specific optimization that tailors pruning configurations to individual prompts, searching for locally optimal patterns that preserve finer visual details and offer more precise prompt-dependent control.
\vspace{-8pt}

As shown in Figure~\ref{fig:single}, the prompt-specific optimization consistently enhances both semantic fidelity and visual coherence compared with dense models and globally optimized CR-Diff. Taking the SDXL $512\times512$ case under Prompt~A as an illustrative example, the dense model on the left fails to express phoenix fire or molten lava and instead resembles a cold carved bird, so the semantic intent is largely lost.
The global CR-Diff result in the middle restores the fiery theme and atmosphere, but the molten quality remains limited.
The prompt-specific optimized result on the right most accurately conveys both the burning phoenix and the flowing rebirth from molten lava, achieving the clearest and most consistent expression of the prompt.

\begin{table*}[t]
\centering
\caption{
Performance comparison on DiTs.
\textbf{Bold} values indicate our CR-Diff is better than the dense model.
The results show that CR-Diff preserves or even improves the generation quality, demonstrating the generalizability of our method beyond UNets.
}

\vspace{-5pt}
\label{table:Dit}
\resizebox{\textwidth}{!}{
\begin{tabular}{l c cc cc cc cc cc}
\toprule
\multirow{2}{*}{Model} & \multirow{2}{*}{Resolution} & \multicolumn{2}{c}{FID $\downarrow$} & \multicolumn{2}{c}{CLIP $\uparrow$} & \multicolumn{2}{c}{ImageReward $\uparrow$}& \multicolumn{2}{c}{PickScore $\uparrow$}& \multicolumn{2}{c}{Aesthetic Score $\uparrow$} \\
\cmidrule(r){3-4} \cmidrule(lr){5-6} \cmidrule(lr){7-8} \cmidrule(lr){9-10} \cmidrule(l){11-12}
& & \multicolumn{1}{c}{Dense} & \multicolumn{1}{c}{CR-Diff} & \multicolumn{1}{c}{Dense} & \multicolumn{1}{c}{CR-Diff} & \multicolumn{1}{c}{Dense} & \multicolumn{1}{c}{CR-Diff} & \multicolumn{1}{c}{Dense} & \multicolumn{1}{c}{CR-Diff} & \multicolumn{1}{c}{Dense} & \multicolumn{1}{c}{CR-Diff} \\
\midrule
SD3Medium & $512 \times 512$ & 40.453 & \textbf{38.901} & 0.317 & \textbf{0.317} & 0.972 & \textbf{1.038} & 22.187 & 22.121 & 4.886 & 4.884 \\
SD3Medium & $1024 \times 1024$ & 37.841 & \textbf{37.026} & 0.320 & \textbf{0.320} & 1.081 & \textbf{1.128} & 22.609 & 22.543 & 5.513 & 5.455 \\
FLUX.dev & $1024 \times 1024$ & 35.799 & \textbf{35.708} & 0.311 & \textbf{0.312} & 0.945 & 0.935 & 22.793 & 22.775 & 6.295 & 6.263 \\

\bottomrule
\end{tabular}
}
\vspace{-8pt}
\end{table*}

\begin{figure*}[t!] 
    \centering 
    \small
    \begin{subtable}[b]{0.68\textwidth} 
        \centering
        \renewcommand{\arraystretch}{1.0}
        \begin{tabular*}{\columnwidth}{@{\extracolsep{\fill}}lcccc}
        \toprule
        \multirow{2}{*}{Model\_Resolution} & \multicolumn{2}{c}{Uniform} & \multicolumn{2}{c}{Block-wise} \\
        \cmidrule(lr){2-3} \cmidrule(lr){4-5}
        & Ratio & IR$\uparrow$ & Ratio & IR$\uparrow$ \\
        \midrule
        SDXL\_$1024\times 1024$ & 0.124 & 0.921 & 0.295 / 0.194 / 0.236 & \textbf{0.946} \\
        SDXL\_$512 \times 512$ & 0.288 & 0.688 & 0.397 / 0.434 / 0.387 & \textbf{0.735} \\
        SD2.1\_$480 \times 360$ & 0.369 & -0.663 & 0.651 / 0.138 / 0.271 & \textbf{-0.561} \\
        \bottomrule
        \end{tabular*}
        \caption{} 
        \label{block-wise-table} 
    \end{subtable}
    \hfill %
    \begin{subfigure}[b]{0.3\textwidth} 
        \centering
        \includegraphics[width=\linewidth]{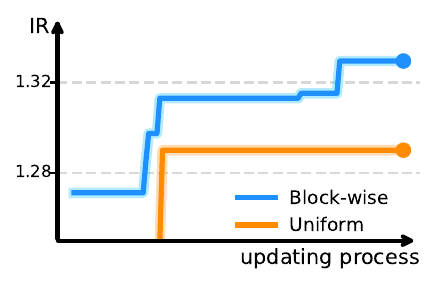}
        \vspace{-20pt}
        \caption{} 
        \label{block-wise-figure} 
    \end{subfigure}
    \vspace{-8pt}
    \caption{Ablation results of block-wise pruning. 
    (a) Performance comparison under uniform and block-wise pruning strategies across different models and resolutions. For block-wise pruning, the ratios are listed in the order \textit{down-sampling / middle / up-sampling}. ImageReward (IR) of the better-performing strategy is highlighted in \textbf{bold}, showing that differentiated ratios improve image quality. 
    (b) Conceptual illustration of ImageReward trends during the updating process for the optimal pruning configuration, showing generally higher and more stable values under block-wise pruning compared to uniform pruning.}
    \label{fig:blockwise} 
\vspace{-13pt}
\end{figure*}

\vspace{0pt}
\subsection{Generalization of CR-Diff}
\label{sec:default and Dit}

\paragraph{Results of CR-Diff on Default Resolutions.}
Although CR-Diff is primarily designed to address degradation at unseen resolutions, it also preserves or even improves model performance at the default resolutions. As shown in Table~\ref{table:AG_default}, across SDXL, SD1.5, and SD2.1, the two-stage framework maintains generative fidelity while often improving metrics. On SDXL at $1024\times1024$, for instance, CR-Diff preserves image fidelity and text–image alignment comparable to the dense model with ImageReward increasing and FID remaining.

\vspace{-8pt}
\paragraph{Results of CR-Diff Applied to DiT.}
Furthermore, while primarily designed for diffusion UNets, CR-Diff can also be safely applied to Diffusion Transformer (DiT) without causing performance degradation. We evaluate CR-Diff on representative DiT models, including SD3Medium~\citep{esser2024scaling} and Flux.dev~\citep{BlackForestLabs2024Flux}. As shown in Table~\ref{table:Dit}, the framework preserves generative fidelity at default resolutions, and in some cases even improves certain metrics such as ImageReward, FID, and CLIP scores. For instance, on SD3Medium at $1024\times1024$, ImageReward increases from 1.081 to 1.128 while FID decreases from 37.841 to 37.026, indicating that CR-Diff's pruning and optimization stages generalize beyond UNet architectures.

\subsection{Ablation Study}
\label{sec:ablation}

\paragraph{Block-Wise Pruning Ratio.}
Table~\ref{block-wise-table} presents representative examples comparing uniform pruning with the proposed block-wise pruning strategy. Across the shown models and resolutions, block-wise pruning yields higher ImageReward scores than uniform pruning. For instance, on SDXL at $1024\times1024$, IR improves from $0.921$ to $0.946$, and on SDXL at $512\times512$, IR improves from $0.688$ to $0.735$. These results reflect the advantage of allocating differentiated pruning ratios that match the functional roles of the corresponding blocks. Full best pruning ratio configurations for all resolution settings are listed in the supplementary material, where substantial differences across downsampling, middle, and upsampling blocks can be observed. 

\begin{figure}
\centering
\captionsetup{font={small}, skip=8pt}
\includegraphics[width=1\linewidth]{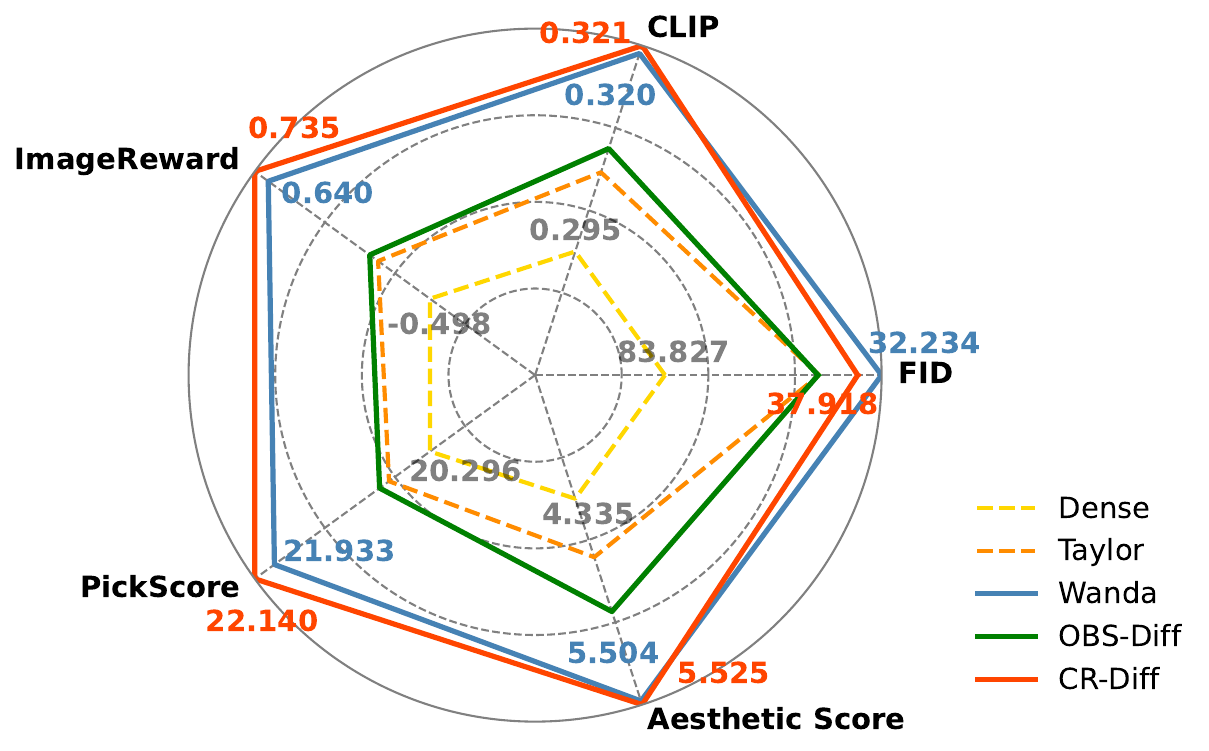}
\vspace{-17pt}
\caption{
Radar comparison across pruning strategies on SDXL at $512\times512$. Five metrics are normalized with direction alignment so that larger radial values denote better performance. CR-Diff achieves the strongest overall results, highlighting its superior perceptual and semantic quality.}
\label{fig:Radar}
\vspace{-14pt}
\end{figure}

The evolution of ImageReward during the optimal pruning config updating process on SDXL $512\times512$ is illustrated in Figure~\ref{block-wise-figure}. Uniform pruning applies the same ratio across all blocks and therefore tends to reduce capacity in regions where parameters are more functionally critical, resulting in a lower and flatter performance plateau during optimization. In contrast, block-wise pruning preserves information flow more effectively, particularly in the middle and upsampling stages that contribute strongly to global structure and fine-grained detail. This leads to a more favorable optimization trajectory and a higher final quality level, as reflected by the consistently stronger ImageReward scores.

\vspace{-12pt}
\paragraph{Comparison of Different Pruning Criteria.}
Unlike traditional pruning methods that mainly aim for efficiency and seek to retain performance comparable to the dense model, CR-Diff is designed to surpass the dense model in generative quality. The magnitude-based pruning in CR-Diff no longer treats gradient magnitude as a sufficient indicator of parameter importance, since a large gradient only reflects strong influence on the output, not whether that influence is beneficial. 

Here we compare our CR-Diff with other representative pruning criteria: Taylor~\citep{molchanov2019importance}, Wanda~\citep{sunsimple}, and OBS-Diff~\citep{zhu2025obs}, on SDXL~\citep{podell2023sdxl} at $512\times 512$. Figure~\ref{fig:Radar} presents the evaluation results of images generated under each pruning strategy alongside the dense model in the form of a radar plot. Although other pruning methods can yield moderate improvements over the dense model, CR-Diff consistently delivers the strongest overall performance across the five metrics. 
Notably, CR-Diff achieves the best results in four of the five metrics.

While Wanda achieves performance relatively close to ours, it requires a full Hessian-free weight importance estimation that takes approximately 420.70s per pruning pass, whereas our magnitude-based block-wise pruning completes in only 0.38s.

\vspace{-5pt}
\begin{table}[t]
\centering
\caption{Ablation results of the pruned output amplification component.
Values in the table denote the performance difference between models with and without POA, where consistent positive values demonstrate the effectiveness of POA.}
\vspace{-8pt}
\label{table:POA_ir}
\small
\setlength{\tabcolsep}{10pt} 
\begin{tabular}{lccc}
\toprule
\multirow{2}{*}{Resolution} & \multicolumn{3}{c}{ImageReward Improvement} \\
\cmidrule(lr){2-4}
& SDXL & SD1.5 & SD2.1 \\
\midrule
$512 \times 512$ & +0.205 & +0.144 & +0.236 \\
$400 \times 560$ & +0.364 & +0.155 & +0.266 \\
$480 \times 360$ & +0.417 & +0.164 & +0.261 \\
\bottomrule
\end{tabular}
\renewcommand{\arraystretch}{1.0}
\setlength{\tabcolsep}{6pt}
\vspace{-12pt}
\end{table}

\paragraph{Pruned Output Amplification.}
Table~\ref{table:POA_ir} highlights the effect of the \textit{pruned output amplification} (POA) mechanism on ImageReward. Across all tested models and resolutions, POA yields consistently positive ImageReward gains over the corresponding pruned baselines. Notably, the improvement becomes even more pronounced under resolution shifts. For instance, on SD1.5, POA yields a +0.144 ImageReward gain at the default $512 \times 512$ resolution, and this improvement further increases to +0.155 and +0.164 at unseen resolutions.
This consistent upward trend suggests that POA serves as an effective component for enhancing pruned diffusion models' performance under cross-resolution conditions.
Full-resolution results and corresponding metrics are deferred to our supplementary material due to limited space, where the aggregated evaluations consistently confirm better performance across all resolutions.

\begin{table}[t]
\centering
\caption{Ablation study on the amplification coefficient $k$ in POA, on SDXL ($512\times512$). A moderate amplification ($k=1.5$) yields the most stable performance gains across evaluation metrics. Best per-metric values are shown in \textbf{bold}. }
\vspace{-8pt}
\label{table:AG_k}
\small
\begin{tabular}{rccc}
\toprule
Metric & K = 1.5 & K = 2.0 & K = 2.5 \\
\midrule
Best IR \% $\uparrow$ & \textbf{54.31} & 29.89 & 15.80 \\ 
\midrule
FID $\downarrow$ & \textbf{37.918} & 43.08 & 61.71 \\ %
CLIP $\uparrow$ & \textbf{0.321} & 0.305 & 0.290 \\ %
ImageReward$ \uparrow$ & \textbf{0.735} & -0.003 & -0.557 \\ %
PickScore $\uparrow$ & \textbf{22.140} & 21.14 & 20.34 \\ 
Aesthetic Score $\uparrow$ & \textbf{5.525} & 5.040 & 4.630 \\ 
\bottomrule
\end{tabular}
\vspace{-12pt}
\end{table}

\vspace{-5pt}
\paragraph{Effect of the Amplification Coefficient $k$.}
To examine the influence of the amplification coefficient $k$ used in pruned output amplification, we conduct an ablation study with $k \in {1.5, 2.0, 2.5}$ and evaluate the resulting generative performance. As shown in Table~\ref{table:AG_k}, $k = 1.5$ yields the most consistent improvements across all metrics, indicating that a moderate amplification effectively strengthens the beneficial deviation introduced by pruning while maintaining coherent semantic structure.

In contrast, increasing $k$ to 2.0 or 2.5 leads to clear degradation. Excessive amplification suppresses meaningful residual signals from the dense output, resulting in weakened semantic alignment and reduced perceptual fidelity. For instance, ImageReward drops from $0.514$ to $-0.557$, and FID rises from $37.21$ to $61.71$ when $k$ increases from 1.5 to 2.5. This highlights that the improvement brought by POA arises from balancing the contributions of the pruned and dense outputs, rather than replacing the latter entirely.

Overall, $k = 1.5$ achieves a stable compromise between preserving semantic faithfulness and enhancing visual quality. Accordingly, we adopt $k = 1.5$ as the default setting in all main experiments.

\vspace{-3pt}

\section{Conclusion}
\label{sec:conclusion}
This work introduces \textit{CR-Diff}, a pruning-based approach to improve cross-resolution consistency in UNet–based text-to-image diffusion models. CR-Diff operates in two stages. First, a \textit{block-wise pruning} strategy allocates differentiated pruning ratios to the downsampling, middle, and upsampling blocks, preserving resolution-stable structure while removing redundant parameters. Second, a \textit{pruned output amplification} mechanism refines the forward denoising trajectory by amplifying the beneficial output tendencies introduced by pruning and suppressing residual artifact-related signals inherited from the dense model.
Unlike existing pruning works that typically pose pruning as an efficiency-improving technique to reduce model size, here we expand its role, \textit{for the first time}, to improving cross-resolution generation quality of diffusion models.
Experiments on SDXL, SD1.5, and SD2.1 demonstrate that CR-Diff enhances perceptual fidelity and semantic coherence at unseen resolutions while preserving performance at default resolutions and on DiT models. CR-Diff also supports optional prompt-specific optimization for adaptive, on-demand enhancement.

\section*{Acknowledgment}
This paper is supported by Young Scientists Fund of the National Natural Science Foundation of China (NSFC) (No. 62506305), and Scientific Research Project of Westlake University (No. WU2025WF003).

{
    \small
    \bibliographystyle{ieeenat_fullname}
    \bibliography{main}
}

\clearpage
\setcounter{page}{1}
\maketitlesupplementary

\section{Block-wise Pruning Ratio Configurations}
\label{sec:ratio_config}
As discussed in Section~\ref{sec:preliminaries}, the UNet architecture comprises downsampling, middle, and upsampling blocks, which differ in redundancy and tolerance to parameter removal. This is further supported by our pruning ratio search experiments across multiple diffusion model families and sampling resolutions, with the resulting block-wise configurations summarized in Table~\ref{tab:ratio_config}.

The empirical results consistently show that the optimal pruning ratios vary across the three block groups in SDXL, SD1.5, and SD2.1, and this difference remains stable when the generation resolution changes. These observations indicate that each block group contributes to synthesis in a structurally differentiated manner and therefore exhibits distinct pruning sensitivity. Applying a uniform pruning ratio across all blocks either disrupts global structural composition or suppresses fine-grained details. In contrast, assigning pruning ratios separately to the downsampling, middle, and upsampling blocks maintains texture fidelity.

Taken together, these findings directly support our \textit{Block-wise Pruning Ratio} Strategy in Section~\ref{sec:blockwise}.

\section{Full Ablation Study of POA}

\label{sec:suppl_POA}
To more comprehensively illustrate the effect of the \textit{pruned output amplification} (POA) mechanism, we provide the full ablation results across models and resolutions in Table~\ref{table:prune_vs_optimized_combined}, which were omitted from the main paper due to space constraints.

This output-level refinement consistently improves generative quality across architectures and resolutions. As shown in Table~\ref{table:prune_vs_optimized_combined}, the refined models achieve stronger semantic consistency and perceptual fidelity, reflected in higher ImageReward and PickScore values compared with pure-pruned baselines. These results indicate that POA functions as a corrective steering mechanism that stabilizes the denoising process and reinforces the desirable generative tendencies of the pruned model while reducing residual artifact-related signals inherited from the dense model.

\vspace{-10pt}
\paragraph{Discussions on Aesthetic Score.}
We observe that Aesthetic Scores may occasionally decrease after applying POA, even as ImageReward, CLIP, and PickScore show consistent improvements. This phenomenon occurs because the Aesthetic Score is particularly sensitive to variations in local texture and stylistic details. By pushing the output further along the pruned direction (k>1), POA naturally moderates certain fine-grained textural components that tend to exhibit instability at unseen resolutions. This moderation results in smoother and more structurally coherent outputs, which may not align perfectly with the Aesthetic Score's emphasis on textural richness. However, the consistent gains observed in ImageReward and PickScore metrics demonstrate improved semantic alignment, enhanced realism, and superior overall visual coherence, thereby validating the effectiveness of POA. 

\begin{table}[t]
\centering
\caption{Block-wise pruning ratio configurations across different models and resolutions, showing the distinct ratio allocations for the downsampling, middle, and upsampling blocks (* indicates the model's default resolution).}
\vspace{-5pt}
\label{tab:ratio_config}

\setcellgapes{3pt} 
\makegapedcells

\begin{tabular*}{\columnwidth}{@{\extracolsep{\fill}}lcc}
\toprule
Model & Resolution & Ratios (Down/Middle/Up) \\
\midrule

\multirow{5}{*}{SDXL} 
& $1024 \times 1024^*$ & 0.295 / 0.194 / 0.236 \\
& $512 \times 512$  & 0.397 / 0.434 / 0.387 \\
& $400 \times 560$  & 0.482 / 0.396 / 0.469 \\
& $480 \times 360$  & 0.434 / 0.428 / 0.355 \\
& $1536 \times 1536$  & 0.300 / 0.343 / 0.300 \\
\midrule

\multirow{4}{*}{SD1.5}
& $512 \times 512^*$  & 0.433 / 0.345 / 0.300 \\
& $400 \times 560$  & 0.319 / 0.240 / 0.192 \\
& $480 \times 360$  & 0.467 / 0.363 / 0.196 \\
& $768 \times 768$  & 0.185 / 0.445 / 0.100 \\
\midrule

\multirow{4}{*}{SD2.1}
& $512 \times 512^*$  & 0.623 / 0.259 / 0.115 \\
& $400 \times 560$  & 0.534 / 0.534 / 0.169 \\
& $480 \times 360$  & 0.651 / 0.138 / 0.271 \\
& $768 \times 768$  & 0.277 / 0.206 / 0.313 \\

\bottomrule
\end{tabular*}
\vspace{-15pt}
\end{table}

\begin{table*}[t]
\centering
\caption{Performance comparison of \textit{pruned} models (the model pruned block-wisely \underline{without} pruned output amplification) and \textit{CR-Diff} (\underline{with} pruned output amplification) across default and unseen resolutions. Resolutions are reported as height $\times$ width, where resolutions marked with * denote the native (default) setting of the model. \textbf{Bold} values indicate that CR-Diff outperforms the pruned baseline. These results verify that POA provides a stable refinement effect that generalizes across models and resolutions.
}
\vspace{-5pt}
\label{table:prune_vs_optimized_combined}
\resizebox{\textwidth}{!}{%
\begin{tabular}{l l cc cc cc cc cc}
\toprule
\multirow{2}{*}{Model} & \multirow{2}{*}{Resolution} & \multicolumn{2}{c}{FID $\downarrow$} & \multicolumn{2}{c}{CLIP $\uparrow$} & \multicolumn{2}{c}{ImageReward $\uparrow$}& \multicolumn{2}{c}{PickScore $\uparrow$}& \multicolumn{2}{c}{Aesthetic Score $\uparrow$} \\
\cmidrule(r){3-4} \cmidrule(lr){5-6} \cmidrule(lr){7-8} \cmidrule(lr){9-10} \cmidrule(l){11-12}
& & \multicolumn{1}{c}{pruned} & \multicolumn{1}{c}{CR-Diff} & \multicolumn{1}{c}{pruned} & \multicolumn{1}{c}{CR-Diff} & \multicolumn{1}{c}{pruned} & \multicolumn{1}{c}{CR-Diff} & \multicolumn{1}{c}{pruned} & \multicolumn{1}{c}{CR-Diff} & \multicolumn{1}{c}{pruned} & \multicolumn{1}{c}{CR-Diff} \\
\midrule
\multirow{5}{*}{SDXL} & $1024 \times 1024$* & 33.397 & 33.562 & 0.322 & 0.322 & 0.834 & \textbf{0.946} & 22.594 & \textbf{22.639} & 6.058 & \textbf{6.106} \\
& $512 \times 512$ & 40.068 & \textbf{37.918} & 0.320 & \textbf{0.321} & 0.530 & \textbf{0.735} & 22.100 & \textbf{22.140} & 5.508 & \textbf{5.525} \\
& $ 400\times560 $ & 43.348 & \textbf{36.688} & 0.308 & \textbf{0.311} & -0.272 & \textbf{0.092} & 20.948 & \textbf{21.074} & 4.752 & 4.672 \\
& $ 480\times360 $ & 56.182 & \textbf{46.040} & 0.301 & \textbf{0.307} & -0.516 & \textbf{-0.099} & 20.636 & \textbf{20.956} & 4.472 & \textbf{4.644} \\
& $ 1536\times1536 $ & 39.362 & 40.380 & 0.312 & 0.312 & 0.108 & \textbf{0.208} & 21.394 & \textbf{21.399} & 5.806 & \textbf{5.855} \\
\midrule
\multirow{4}{*}{SD1.5} & $ 512\times512 $* & 39.563 & \textbf{37.773} & 0.313 & \textbf{0.314} & 0.059 & \textbf{0.203} & 21.376 & \textbf{21.377} & 5.265 & 5.233 \\
& $ 400\times560 $ & 40.188 & \textbf{39.291} & 0.309 & \textbf{0.310} & -0.004 & \textbf{0.151} & 21.143 & \textbf{21.188} & 4.785 & 4.779 \\
& $ 480\times360 $ & 39.774 & \textbf{37.634} & 0.305 & \textbf{0.307} & -0.190 & \textbf{-0.026} & 20.931 & \textbf{20.944} & 4.848 & 4.819 \\
& $ 768\times768 $ & 39.084 & \textbf{38.452} & 0.314 & \textbf{0.315} & -0.063 & \textbf{0.059} & 21.190 & \textbf{21.232} & 5.389 & 5.385 \\
\midrule
\multirow{4}{*}{SD2.1} & $ 512\times512 $* & 38.799 & \textbf{36.792} & 0.306 & \textbf{0.309} & -0.288 & \textbf{-0.052} & 20.940 & \textbf{20.960} & 5.174 & 5.082 \\
& $ 400\times560 $ & 38.344 & \textbf{35.837} & 0.301 & \textbf{0.304} & -0.334 & \textbf{-0.068} & 20.565 & 20.540 & 4.559 & 4.428 \\
& $ 480\times360 $ & 43.294 & \textbf{41.042} & 0.290 & \textbf{0.294} & -0.822 & \textbf{-0.561} & 20.090 & \textbf{20.177} & 4.564 & 4.532 \\
& $ 768\times768 $ & 35.595 & \textbf{35.237} & 0.317 & \textbf{0.318} & 0.304 & \textbf{0.419} & 21.497 & 21.451 & 5.429 & 5.339 \\
\bottomrule
\end{tabular}%
}
\vspace{-10pt}
\end{table*}

\section{Simulated Annealing (SA) Algorithm}

Algorithm~\ref{alg:sa} summarizes the simulated annealing (SA) routine used to search for the optimal pruning ratio configuration $\mathbf{r} = { r_{\text{down}}, r_{\text{mid}}, r_{\text{up}} }$. The hyperparameters include the initial temperature $T_{init}$, cooling rate $\alpha$, iteration budget $N_{iter}$, a set of candidate seeds $S_{seeds}$, and a restart limit $R_{max}$. Starting from the best candidate in the initial seed set, the algorithm iteratively samples neighboring configurations and accepts them based on the standard SA criterion, allowing occasional uphill moves to escape local minima. A lightweight reheating and restart mechanism is incorporated to prevent stagnation and maintain exploration when the search plateaus. This SA variant provides a simple and robust way to obtain near-optimal ratio configurations without exhaustive search, and the resulting best state $S_{best}$ serves directly as the optimal pruning–ratio configuration $\mathbf{r}$.

\section{Analyses on Unseen Resolutions}
\begin{table*}[t]
\centering
\caption{
Evaluation of SDXL at the unseen higher resolution $1536\times1536$. The results show that the model remains stable at this scale and CR-Diff maintains comparable performance without altering semantic structure or perceptual characteristics.
}
\vspace{-5pt}
\label{table:sdxl_1536}
\resizebox{\textwidth}{!}{%
 \begin{tabular}{l c cc cc cc cc cc}
 \toprule
 \multirow{2}{*}{Model} & \multirow{2}{*}{Resolution} & \multicolumn{2}{c}{FID $\downarrow$} & \multicolumn{2}{c}{CLIP $\uparrow$} & \multicolumn{2}{c}{ImageReward $\uparrow$}& \multicolumn{2}{c}{PickScore $\uparrow$}& \multicolumn{2}{c}{Aesthetic Score $\uparrow$} \\
 \cmidrule(r){3-4} \cmidrule(lr){5-6} \cmidrule(lr){7-8} \cmidrule(lr){9-10} \cmidrule(l){11-12}
 & & \multicolumn{1}{c}{dense} & \multicolumn{1}{c}{ours} & \multicolumn{1}{c}{dense} & \multicolumn{1}{c}{ours} & \multicolumn{1}{c}{dense} & \multicolumn{1}{c}{ours} & \multicolumn{1}{c}{dense} & \multicolumn{1}{c}{ours} & \multicolumn{1}{c}{dense} & \multicolumn{1}{c}{ours} \\
 \midrule
SDXL& $ 1536\times1536 $ & 46.563 & 40.380 & 0.315 & 0.312 & 0.300 & 0.208 & 21.675 & 21.399 & 5.952 & 5.855 \\
\bottomrule
\end{tabular}%
}
\end{table*}

Beyond the detailed analysis in Section~\ref{sec:unseen}, which demonstrates consistent improvements under CR-Diff at unseen resolutions, we provide additional analyses at higher resolutions for SDXL.
SDXL, natively trained at $1024\times1024$ with a resampler and high-resolution cross-attention, effectively internalizes dense object structures and sharp boundaries. As a result, scaling to $1536\times1536$ does not lead to noticeable degradation, with FID remaining low and perceptual metrics such as CLIP, PickScore, and Aesthetic Score staying stable as shown in Table~\ref{table:sdxl_1536}. Notably, under this higher resolution, pruning-based CR-Diff successfully preserves SDXL's original generative characteristics.
\setlength{\textfloatsep}{5pt}
\begin{algorithm}[t]
\caption{Simulated Annealing for the Optimal Pruning Ratio Configuration $\mathbf{r}$}
\label{alg:sa}
\KwData{$T_{init}, \alpha, N_{iter}, S_{seeds}, R_{max}$}
\KwResult{$S_{best}$}

$(S_{curr}, E_{curr}) \leftarrow \text{BestSeed}(S_{seeds})$\;
$S_{best} \leftarrow S_{curr}$; $E_{best} \leftarrow E_{curr}$\;
$T \leftarrow T_{init}$; $C_{restart} \leftarrow 0$\;

\For{$i = 1$ \textbf{to} $N_{iter}$}{
    $S_{neighbor} \leftarrow \text{GenerateNeighbor}(S_{curr}, T)$\;
    $E_{neighbor} \leftarrow \text{Evaluate}(S_{neighbor})$\;

    \If{$E_{neighbor} < E_{curr}$ \textbf{or} 
        $\exp(-(E_{neighbor}-E_{curr})/T) > \text{rand}(0,1)$}{
        $S_{curr} \leftarrow S_{neighbor}$\;
        $E_{curr} \leftarrow E_{neighbor}$\;
    }

    \If{$E_{curr} < E_{best}$}{
        $S_{best} \leftarrow S_{curr}$\;
        $E_{best} \leftarrow E_{curr}$\;
    }

    $T \leftarrow \alpha \cdot T$\;

    \If{$T$ is too small}{
        $T \leftarrow T_{init}$; 
        $C_{restart} \leftarrow C_{restart} + 1$\;
        \If{$C_{restart} > R_{max}$}{\textbf{break}}
    }
}
\Return $S_{best}$\;

\end{algorithm}

\section{Expanded Qualitative Analyses}
\begin{figure*}
\centering
\captionsetup{font={small}, skip=8pt}
\includegraphics[width=1\linewidth]{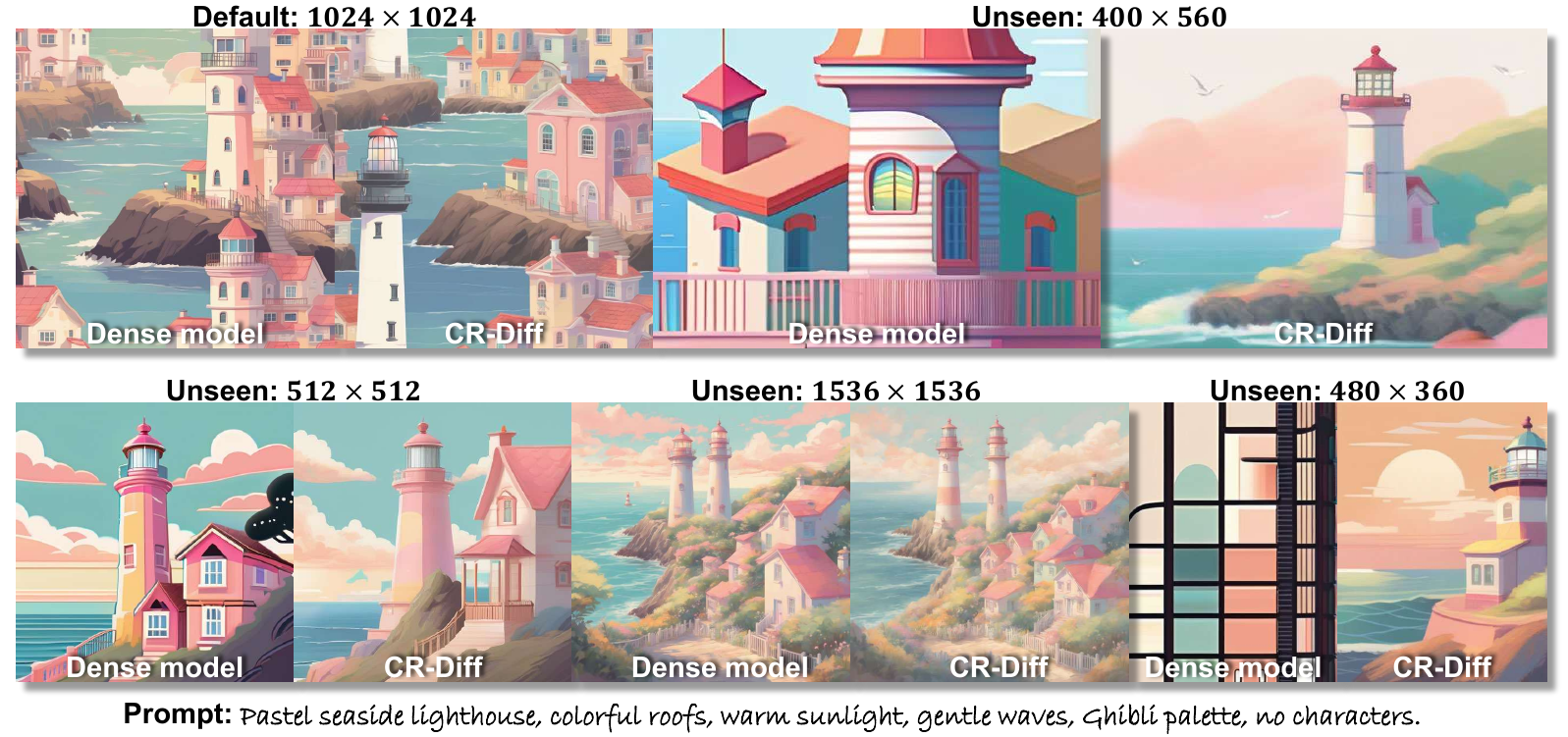}
\vspace{-15pt}
\caption{Additional cross-resolution comparisons between SDXL~\citep{podell2023sdxl} and its CR-Diff counterpart. CR-Diff consistently improves cross-resolution image quality compared to the dense model.}
\label{fig:lighthouse}
\end{figure*}

\begin{figure*}
\centering
\captionsetup{font={small}, skip=8pt}
\includegraphics[width=1\linewidth]{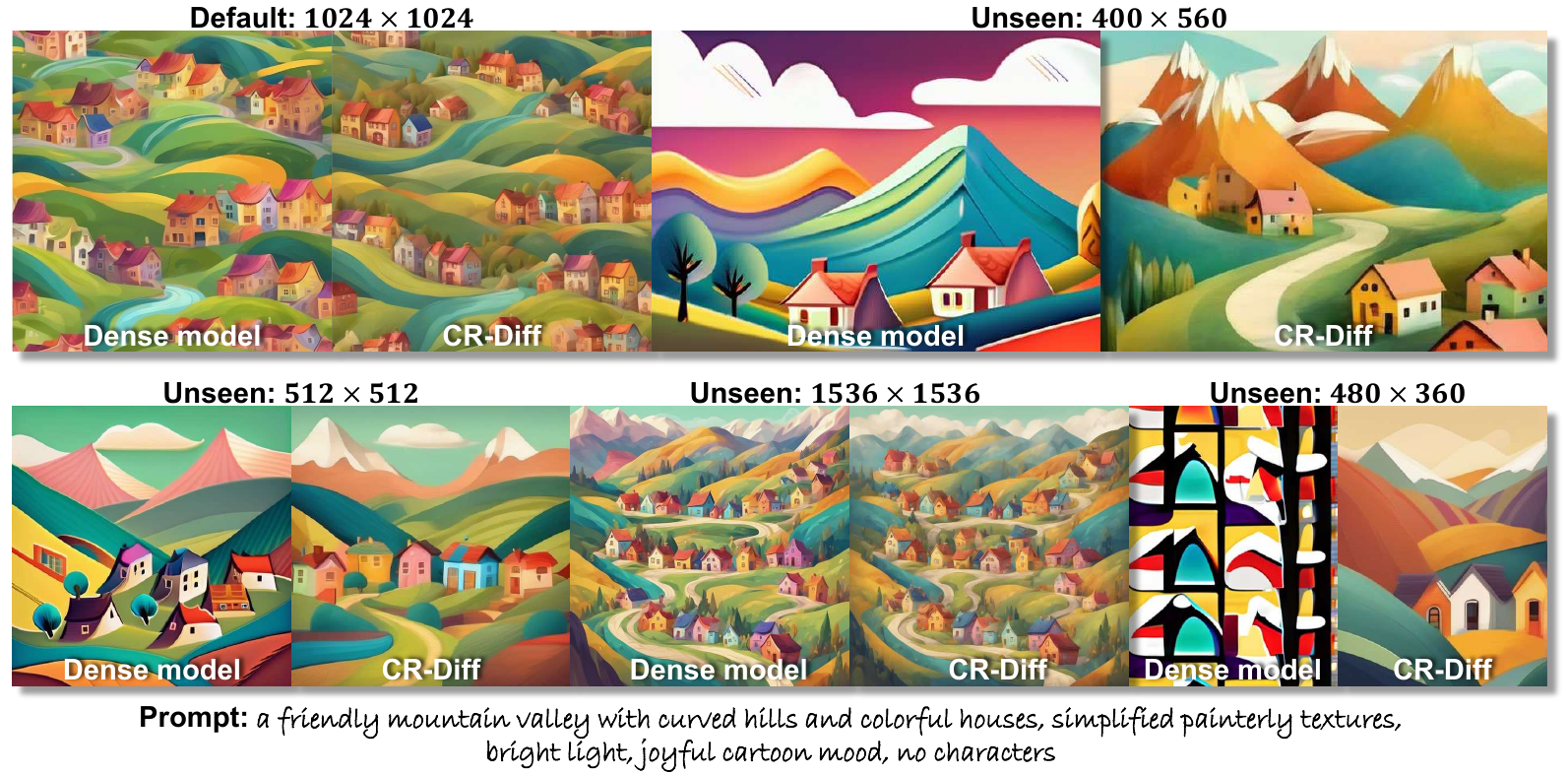}
\vspace{-15pt}
\caption{Additional cross-resolution comparisons between SDXL~\citep{podell2023sdxl} and its CR-Diff counterpart. CR-Diff consistently improves cross-resolution image quality compared to the dense model.}
\label{fig:valley}
\end{figure*}

\paragraph{Representative Teaser Results.}
In Figures~\ref{fig:lighthouse} and \ref{fig:valley}, we present additional representative teaser examples following the style of Figure~\ref{fig:teaser}, further illustrating the effectiveness of CR-Diff in enhancing cross-resolution visual consistency over the dense SDXL~\citep{podell2023sdxl}.

\begin{figure*}
\centering
\captionsetup{font={small}, skip=8pt}
\includegraphics[width=1\linewidth]{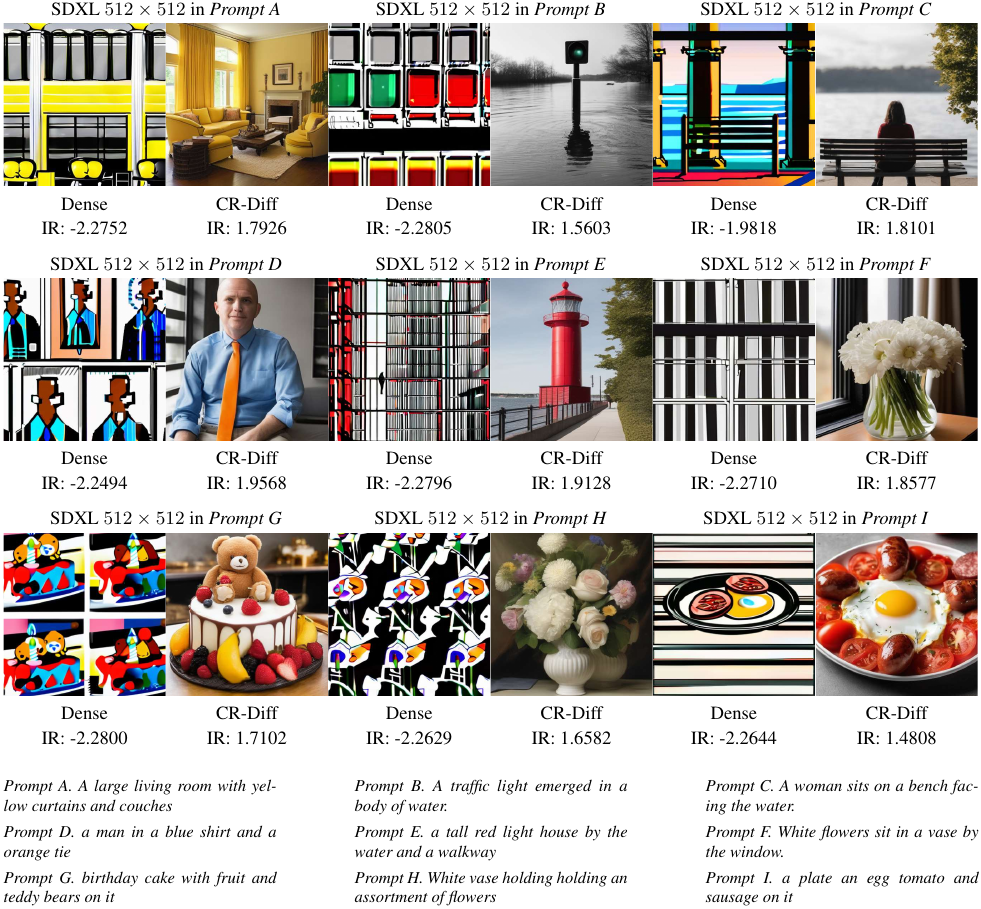}
\vspace{-15pt}
\caption{Additional cross-resolution comparison on a subset of 5K prompts from the MS-COCO 2014 validation set~\citep{lin2014microsoft}. CR-Diff shows consistent gains in both ImageReward and visual fidelity compared to the original SDXL. \textit{Dense} denotes the original unpruned model. Each group corresponds to a specific prompt, and the ImageReward (IR) scores are shown below each image.}
    \label{fig:5k_sdxl}
\end{figure*}

\begin{figure*}
\centering
\captionsetup{font={small}, skip=8pt}
\includegraphics[width=1\linewidth]{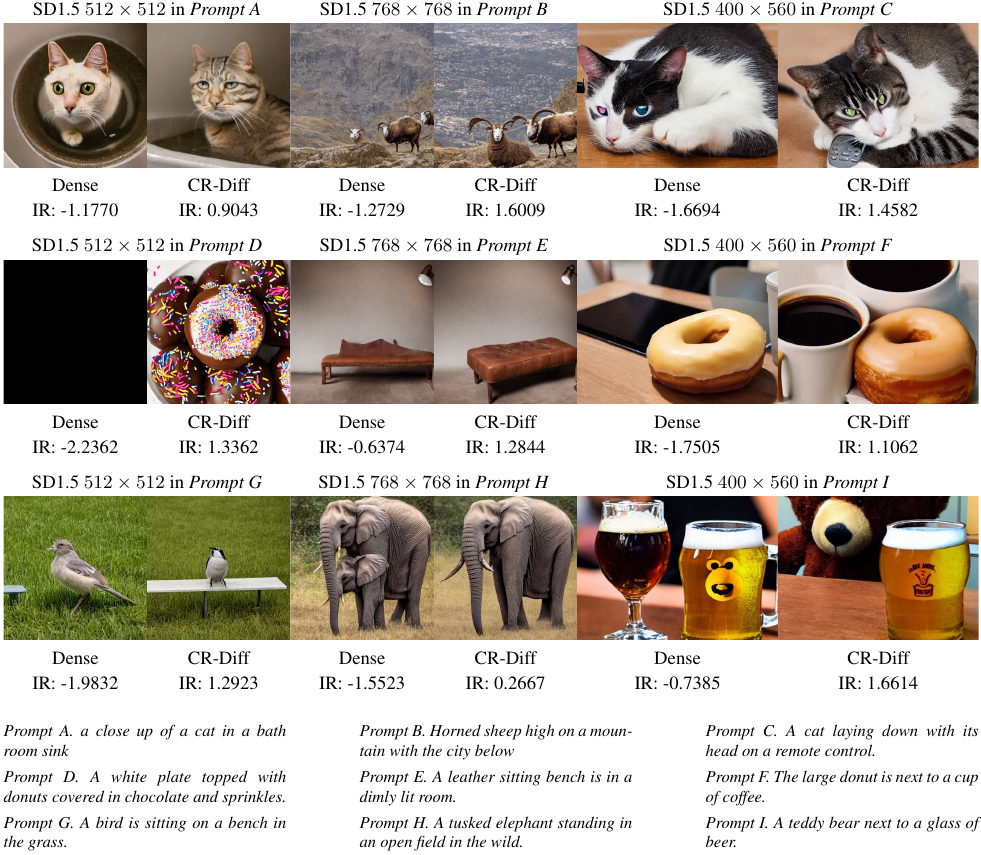}
\vspace{-15pt}
\caption{Additional cross-resolution comparison on a subset of 5K prompts from the MS-COCO 2014 validation set~\citep{lin2014microsoft}. CR-Diff shows consistent gains in both ImageReward and visual fidelity compared to the original SD1.5. \textit{Dense} denotes the original unpruned model. Each group corresponds to a specific prompt, and the ImageReward (IR) scores are shown below each image.
}
    \label{fig:5k_sd1.5}
\end{figure*}

\begin{figure*}
\centering
\captionsetup{font={small}, skip=8pt}
\includegraphics[width=1\linewidth]{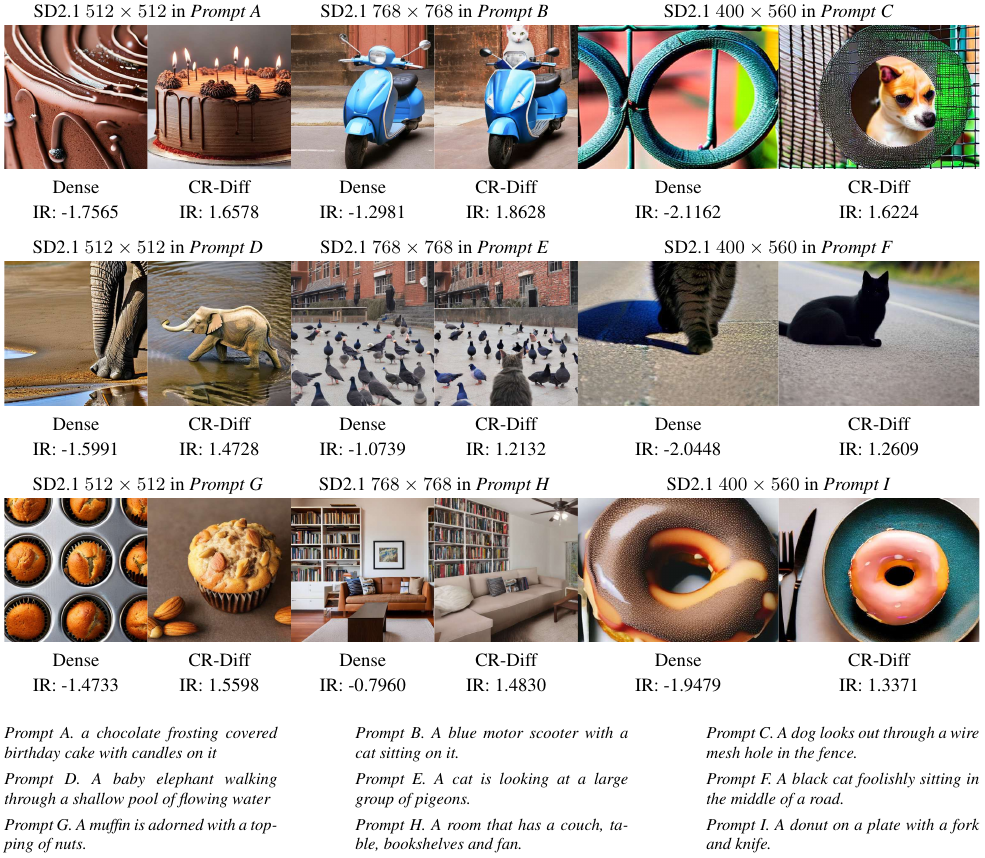}
\vspace{-15pt}
\caption{Additional cross-resolution comparison on a subset of 5K prompts from the MS-COCO 2014 validation set~\citep{lin2014microsoft}. CR-Diff shows consistent gains in both ImageReward and visual fidelity compared to the original SD2.1. \textit{Dense} denotes the original unpruned model. Each group corresponds to a specific prompt, and the ImageReward (IR) scores are shown below each image.
}
    \label{fig:5k_sd2.1}
\end{figure*}

\vspace{-10pt}
\paragraph{Results on the 5K Dataset.}
In Figures~\ref{fig:5k_sdxl}, \ref{fig:5k_sd1.5}, and \ref{fig:5k_sd2.1}, we present additional results on a subset of 5K prompts sampled from the MS-COCO 2014 validation set~\citep{lin2014microsoft}, evaluated with SDXL, SD~2.1, and SD~1.5 across multiple resolutions. These examples show clear improvements in ImageReward and exhibit noticeably better structure preservation, semantic consistency, and fine-grained visual fidelity.

\begin{figure*}
\centering
\captionsetup{font={small}, skip=8pt}
\includegraphics[width=1\linewidth]{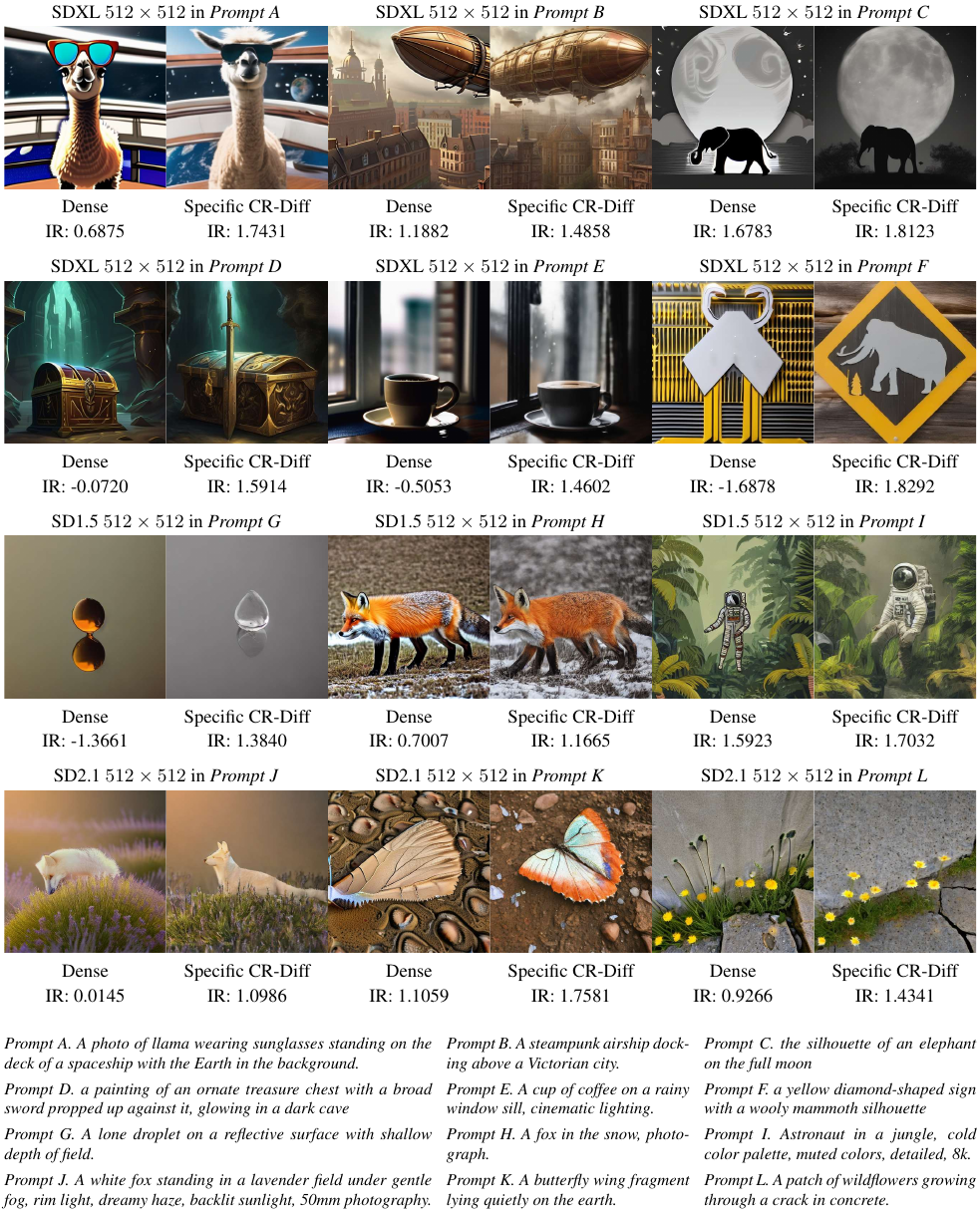}
\vspace{-15pt}
\caption{Visual comparison across two generation settings. 
    \textit{Dense} denotes the original unpruned model. 
    \textit{Specific CR-Diff} adjusts the pruning ratio based on each input prompt, enabling prompt-tailored optimization.
    Each group corresponds to a specific prompt, and the ImageReward (IR) scores are shown below each image. 
  Both quantitative and qualitative results show that Specific CR-Diff improves semantic alignment and visual coherence for the given prompt.
}
    \label{fig:suppl_single}
    \vspace{-5pt}
\end{figure*}

\vspace{-10pt}
\paragraph{Extended Results for Prompt-Specific Optimization.}
In Figure~\ref{fig:suppl_single}, we present extended qualitative results 
from our prompt-specific optimization mentioned in 
Section~\ref{sec:single}, highlighting clear improvements in ImageReward and stronger prompt–detail correspondence across diverse input prompts.

\end{document}